\documentclass[10pt,letterpaper]{article}
\date{}

\usepackage{csquotes} 
\usepackage{natbib}
\usepackage{times}  %
\usepackage{helvet}  %
\usepackage{courier}  %
\usepackage{url}  %
\usepackage{graphicx}  %
\usepackage{subfigure}
\usepackage{amsmath}
\usepackage{amssymb}
\usepackage{color}
\usepackage{tikz}
\usepackage{float}
\usepackage[font=small,labelfont=bf]{caption}

\newcommand{\tikzdrawcircle}[2][black,fill=black]{\tikz[baseline=-0.5ex]\draw[#1,radius=#2] (0,0) circle ;}%

\makeatletter
\let\@fnsymbol\@arabic
\makeatother

\begin{document}
  	\title{\bf What Stands-in for a Missing Tool? \\ A Prototypical Grounded Knowledge-based Approach to Tool Substitution}
  	\author{Madhura Thosar\thanks{Madhura Thosar and Sebastian Zug are with Faculty of Computer Science, Otto-von-Guericke University Magdeburg, Germany, {\tt\small thosar@iks.cs.ovgu.de, zug@ivs.cs.uni-magdeburg.de}}\ , Christian A. Mueller\thanks{Christian A. Mueller is with the Robotics Group, Computer Science \& Electrical Engineering Department, Jacobs University Bremen, Germany, {\tt\small chr.mueller@jacobs-university.de}}\ , Sebastian Zug\footnotemark[1]}
  	\maketitle
\begin{abstract}
When a robot is operating in a dynamic environment, it cannot be assumed that a tool required to solve a given task will always be available. 
In case of a missing tool, an ideal response would be to find a substitute to complete the task.
In this paper, we present a proof of concept of a grounded knowledge-based approach to tool substitution.
In order to validate the suitability of a substitute, we conducted experiments involving $22$ substitution scenarios.
The substitutes computed by the proposed approach were validated on the basis of the experts' choices for each scenario.
Our evaluation showed, in $20$ out of $22$ scenarios ($91\%$), the approach identified the same substitutes as experts.
\end{abstract}

\section{Introduction}
\label{sec:introduction}

The sophistication pertaining to tool-use in humans involves not just the dexterity in manipulating a tool, but also the diversity in tool exploitation.
The ability to exploit the tools has enabled humans to adapt and thus exert control over an uncertain environment, especially when they are faced with unfavorable situations.
For example, if we don't find a hammer to hammer a nail into a wall, we will use a heel of a shoe or a rock or if a tray is unavailable for serving the drinks, we will use a plate for serving.
In situations like these, humans seem to know - either from the past experience or from observations or from the ``necessity is the mother of improvisation (invention)'' type approach - what \emph{kind} of object is needed as a substitute.

On the contrary, consider a robot performing a task that involves tool use.
When a robot is operating in a dynamic environment, it can not be assumed that a tool required in the task will always be available. 
In situations like these, an effective way for a robot would be to find an alternative as humans do, for example, use an eating plate for serving, rather than wait until a tray becomes available. 
This skill is significant when operating in a dynamic, uncertain environment because it allows a robot to adapt to unforeseen situations to a degree.
The question is how can a robot determine which object in the environment is a viable candidate for a substitute?
A possible approach would be by interacting with an object in a manner missing tool is maneuvered.
However, it would be time consuming if a robot interacts with every single object in the environment to determine a viability which makes this approach less practical.

In this prototypical work, we propose a non-invasive approach that identifies viable candidate/s from the existing objects in the environment.
This paper makes the following contributions:
1) An approach to create grounded knowledge about objects expressed in terms of their properties (Sec. \ref{subsec:KB_method}), 
2) %
an approach to identify relevant properties of a missing tool and determine a substitute on the basis of them (Sec. \ref{subsec:reasoner}). %

\section{Related Work}
\label{sec:related_work}
Typically, a substitute for a missing tool is determined by means of knowledge base that provides knowledge about objects and similarity measures to determine the similarity between a missing tool and a potential substitute. 
In the following, in addition to the approaches to determine a substitute, we also report the literature related to existing knowledge bases developed for robotic applications.

\subsubsection{Knowledge Base}
\label{subsec:related_kb}

We reviewed in \cite{Thosar2018} nine existing knowledge bases namely: KNOWROB \cite{Tenorth}, MLN-KB \cite{Zhu2014}, NMKB \cite{Pineda2017}, OMICS \cite{Gupta2004}, OMRKF \cite{Suh2007}, ORO \cite{Lemaignan2010}, OUR-K \cite{Lim2011}, PEIS-KB \cite{Daoutis2009}, and RoboBrain \cite{Saxena2014}.
The objective was to determine whether these existing knowledge bases contain 1) ontological knowledge about the properties of objects, 2) such knowledge is grounded into robot's perception, and 3) intra-class variability in a property is modeled instead of expressing the property in a binary form.
We gained primarily the following insights which form the basis for our work.

We noted that the majority of the knowledge bases relied on the external human-centric commonsense knowledge bases such as WordNet, %
Cyc, %
OpenCyc, %
and some either relied on the hand-coded knowledge %
or on the knowledge acquired by human-robot interaction. %
The main issue, we believe is that, the depth and breadth of the human-centric knowledge base is not observable by a robot in its entirety due to its limited sensing capabilities. 
This causes a disconnect between human-centric knowledge and robot-centric perception.
To deflect this issue, we aim to acquire the robot-centric perceptual data for different properties of objects.
Such property data can then be used to generate grounded knowledge about objects (see Sec. \ref{subsec:know_acqui}).

\subsubsection{Substitution Computation}
\label{subsec:related_sub_comp}

One of the closest areas that study the usability of an object is affordancs of tools where the primary focus is to examine various functional abilities of an object by exploring what actions can be performed on the object and observing its responses.
As such, using a substitute in place of a missing tool can also be seen as transferring of an affordance of the missing tool to the substitute after determining similarity between them.%

In \cite{Awaad2014a}, a substitute for a missing tool is inferred on the basis of inheritance and equivalence relations.
The work discussed in \cite{Agostini2015} retrieves the knowledge about objects from the ROAR \cite{Szedmak2014} relational database and determines a substitute that shares similar affordances.
However, in ROAR, the knowledge is acquired either using machine learning techniques requiring training examples or inferred or hand-coded.
The work in \cite{Boteanu2016} uses the ConceptNet where potential candidates are extracted from the knowledge base if they share the same parent with a missing tool for the predetermined relations: \textit{has-property}, \textit{capable-of} and \textit{used-for}.
After eliminating irrelevant candidates, a substitute is determined on the basis of the similarity metrics.
The approach proposed in \cite{Abelha2016} uses a part-based 3D model and weight of an object to determine the orientation and manipulation of a substitute to be used as a missing tool.
In the cases where supervised machine learning technique is used, providing bulk of labeled examples beforehand would not be realistic for a substitution problem scenario.
On the other hand, the approaches which rely on existing external knowledge bases are built around the available knowledge in the knowledge bases which does impose some constraints. %
We circumvents this issue by first identifying what knowledge is generally required to determine a substitute and then build an approach to acquire the required knowledge and compute a substitute on the basis of it.

\section{Challenges}
\label{sec:challenges}

\textit{How to characterize similarity between a missing tool and a potential substitute: } 
A candidate for a substitute is expected to be similar to a missing tool to some degree to ensure a substitutability.
The notion of similarity can be understood in various forms, for instance, a distance between two objects denoted by two points in a multi-dimensional space or two objects belonging to the same cluster or aspects of the objects that are identified as shared. 
In this work, the question will be addressed in a broader sense: it is not merely about identifying a similar object by deploying some similarity measure, instead, it is about gaining an access to what aspects of the objects were found to be shared between the similar objects.

\textit{What kind of knowledge is required to determine the similarity: } 
It has been demonstrated in the literature on tool use in humans and animals alike that in order to use an object in tasks one needs to have knowledge about objects \cite{Baber2003}.
Baber in \cite{Baber2003_1} also noted that conceptual knowledge about objects is especially desired in tool use where a systematic deliberation is called for.
For a robot, the story won't be much different if it is expected to perform in the real world along side humans.
As a consequence, a robot needs conceptual knowledge about an object where the object will not be only a physical entity that is merely to be perceived, but also a concept which consists of distinct characteristics and relations which set each object apart from each other and also similar to each other.

\textit{How to acquire the necessary knowledge: }
The acquisition of such conceptual knowledge is not without challenge.
From a robot stand-point, it is a trade-off between what needs to be known and what can be known.
The trade-off is a direct consequence of the limited perception capabilities of a robot which often leads to partial understanding of the environment.
While deploying a multi-modal perception to extract the required knowledge about objects would be an ideal solution, however, it carries its own set of complexities such as noisy sensors, dynamicity of the environment, complexities of the composition of an object.
For this prototypical work, the necessary knowledge is acquired using human-centric as well as machine-centric methods.

\textit{How to maneuver a substitute as a missing tool: }
Once the substitute has been identified, a robot is expected to use it in place of a missing tool and achieve the same result as the missing tool in the task.
The challenge to estimate the maneuver as well as grasping of a substitute is two fold: to determine whether the maneuver and grasping knowledge of a missing tool can be transferred and utilized on a substitute, else, estimate the maneuver and grasping for a substitute such that it can be used as a missing tool in the task.

For this work, we have focused on the first three challenges and have developed a prototypical system called ERSATZ (German word for a substitute or alternative) where the focus is to identify the required knowledge to determine a substitute and develop a system that computes a substitute for a missing tool.

\section{Approach}
\label{sec:approach}

The proposed approach distinguishes a tool from a substitute where a tool is defined as an artifact that is designed, manufactured and maneuvered in accordance with its designated purpose in the tasks such as hammer for hammering, tray for serving etc., while a substitute is seen as an extension of a missing tool.
Within the context of a designated purpose, the relationship between a tool and a substitute is symmetric, for instance, for hammering, a hammer can be replaced by a heeled shoe and vice versa.
However, it may not always be the case once you step outside the context, for instance, a hammer can not replace a heeled shoe.
Our research work, therefore, focuses on searching for a substitute for a conventional tool required in the ongoing task as opposed to determining a substitute for itself.

Consider a scenario in which a robot has to choose between a plate and a mouse pad as an alternative for a tray.
A tray can be defined as a rigid, rectangular, flat, wooden, brown colored object while a plate can be defined as a rigid, circular, semi-flat, white colored object and a mouse pad as soft, rectangular, flat, leather-based object.
Bear in mind, however, that some properties are more relevant than others with respect to the designated purpose of the tool.
For a tray whose designated purpose is \textit{to carry}, rigid and flat are more relevant to \textit{carry} than a material or a color of a tray.
Consequently, to find the most appropriate substitute, the relevant properties of the unavailable tool need to correspond to as large a degree as possible to the properties of the possible choices for a substitute. 

The proposed approach performs conceptual knowledge-driven computation to identify the relevant properties of the missing tool and determines the most similar substitute on the basis of those properties.
Besides identifying the most similar object as a substitute for a missing tool, the proposed approach grants an explicit access to the relevant properties of the missing tool which carries twofold advantages: firstly, knowing which properties are primarily required in the potential substitute narrows down the search space and secondly, in case of an unknown object instance, only the relevant properties will have to be learned to determine a substitute.

The conceptual knowledge considered in this work primarily involves properties of the objects.
The properties considered are divided into \emph{physical} and \emph{functional} properties where physical properties describe the physicality of the objects such as \emph{rigidity}, \emph{weight}, \emph{hollowness} while the functional properties ascribe the (functional) abilities or affordances to the objects such as \emph{containment}, \emph{blockage}, \emph{support}.
The functional properties in the proposed approach play a primary role in identifying the relevant properties of a missing tool (see Sec. \ref{subsec:reasoner}).

The functional properties considered in this work are derived from the theory of image schemas \cite{Gardenfors1987} which has its roots in cognitive linguistics. 
According to \cite{Kuhn2007} \textit{image schemas are patterns abstracted from spatio-temporal experiences.}
Essentially image schemas capture recurrent patterns that emerge from our perceptual and bodily interactions with the environment. 
Since some of these patterns are posited on the operational abilities of objects Kuhn postulated in \cite{Kuhn2007} that affordances a.k.a. functional properties \cite{Gardenfors1987} for the spatio-temporal processes can be derived from image schemas. 
For example, the \textit{containment} schema suggest an object's ability to contain something or the \textit{support} schema indicates an object's ability to hold up something or the \textit{blockage} refers to the ability of an object to block or obstruct the movement of an other object.
Currently, the proposed system is restricted to three functional properties based on image schemas: \textit{containment}, \textit{support} and \textit{blockage}.
While the functional properties as well as the designated purpose can both be identified as affordances, the proposed approach is built by hypothesizing that the functional properties are building blocks upon which designated purposes of tools rest.

\section{Methodology}
\label{sec:method}
\subsection{Knowledge Acquisition}
\label{subsec:know_acqui}
Our ultimate objective is to acquire machine centric data from which property specific data can be extracted.
Such property data will then be used to generate grounded knowledge about objects.
As a first step, our initial property acquisition focuses on the composite of a machine-centric and a human-centric method.
In the machine-centric approach, geometrical properties are acquired using a non-invasive vision-based technique 
while non-geometric properties are acquired by sampling from the data from the expert generated intuitive model for the properties.

\textbf{Machine Generated Properties:}
In this paper, we introduce
a state-of-art data-driven approach that unsupervisedly conceptualizes shape according to commonalities within object point clouds which is discussed in detail in  our work \cite{mueller_iros_2018}.
As a result of the process, a set of shape concepts is generated which concept responses for an unknown object are used in the knowledge base as machine-generated geometric object properties.

\begin{figure}[tb]
	\small
	\centering
	\includegraphics[width=0.99\linewidth]{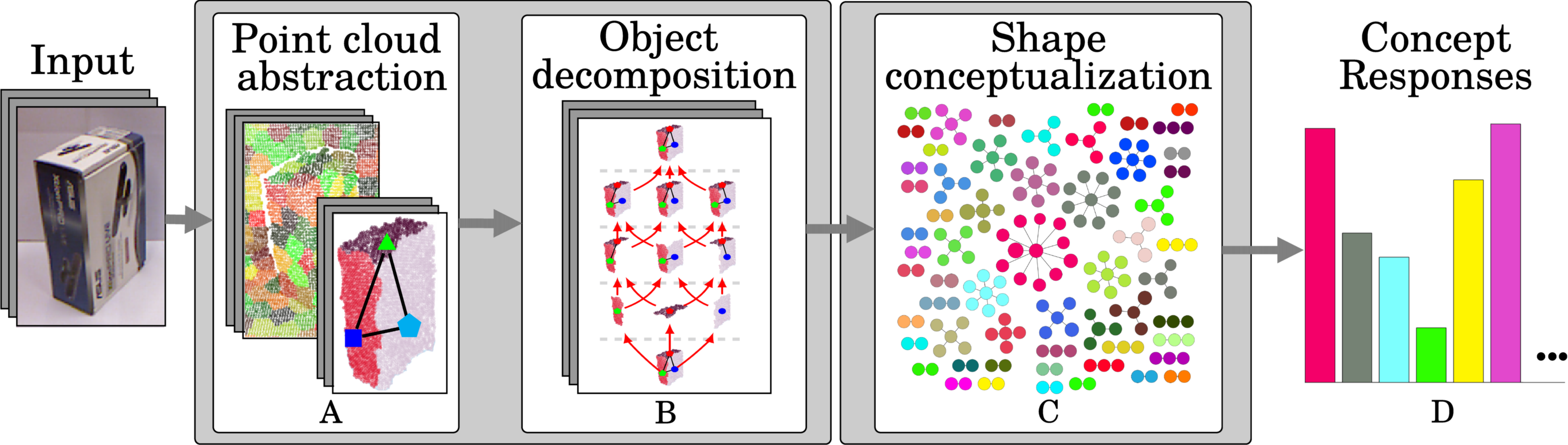}
	\caption{Illustration of the object shape conceptualization approach~\cite{mueller_iros_2018}. Concepts are randomly colored.}
	\label{fig:approach_illustration}
\end{figure}

In our previous work on shape concept learning~\cite{mueller_iros_2018}, raw sensor information in form of point clouds is abstracted to a symbolic level in which point cloud segments~\cite{MuellerBirkIcra2016} may represent meaningful shape components in a symbolic space~\cite{6942984}.
Therein we introduce a hierarchical learning procedure that leads to symbols which are gradually organized to reflect generic-to-specific facets of shape components and can be subsequently used as building blocks that constitute objects (see A in Fig.~\ref{fig:approach_illustration}).
 
An object shape representation is introduced that gradually encodes observed objects symbol compositions (see B in Fig.~\ref{fig:approach_illustration}): from local components to component groups that may represent object parts or objects as a whole.
The proposed shape representation incorporates aspects of exemplar, respectively, prototype theory since we believe that the richness of a prototype provides an unaltered perspective on the characteristics of object instances.
Based on the proposed symbolic shape representation we analyze \emph{topology} and \emph{structure} within the encoded symbol \emph{compositions} in order to discover persistent patterns that may represent \emph{shape concepts}.

We introduce an iterative filtering process~\cite{mueller_iros_2018} to associated instances to groups which may represent shape concepts (see C in Fig.~\ref{fig:approach_illustration}). 
Given the set of learned concepts, for an unknown object, concept responses are retrieved (see D in Fig.~\ref{fig:approach_illustration}) and exploited as machine-generated geometric object property values in our tool-substitution scenario. 
Note that in our tool-substitution scenario, concepts are learned from unlabeled object instances of the \emph{Object Discovery Dataset(ODD)}~\cite{MuellerBirkIcra2016}; the ODD provides a variety of objects from \emph{teddy bears} over \emph{flash lights} to \emph{shoes} which facilitates an expressive concept generation.

\textbf{Human Generated Properties}
The geometric properties alone offer a very limited scope of the physicality as well as the functionality of an object. Therefore, to compensate the gap, we also considered non-geometrical properties such as weight, rigid, hollowness as physical and support, blockage, containment as functional.
Note that, in general, these properties are challenging and cumbersome to extract solely from non-invasive visuoperceptual approaches.
Consequently, extracting such properties via multi-modal or manipulation capabilities is needed, but this is beyond the scope of this paper. 
In the generation process, a set of labeled prototype objects selected from the Washington dataset (see Table \ref{tab:datasize}) were taken into account.
The distribution of each property for particular object labels (cf. Table~\ref{tab:datasize}) was approximated by an expert to resemble the scope for the variations in the values of the property in general.
Consequently, given an object and its label, a sample value was drawn from the a-priori generated property distribution.

\subsection{Knowledge about Objects}
\label{subsec:KB_method}
Knowledge about objects is spread across three levels: the first level consists of the data about the machine-generated as well as human-generated properties, the second level consists of qualitative knowledge about individual object instances, while the third level consists of the aggregated qualitative fuzzy knowledge about respective classes of object instances.
The fuzzy formalism is used to model the intra-class variations in the objects.
In the following, we discuss the formal description of the methodology deployed to create grounded knowledge about objects. 

Consider \textbf{O} as a given set of object class labels where (by abuse of notation) each object class is identified with its label.
Let each object class $O \in \textbf{O}$ be a given set of its instances. 
Let $\bigcup \textbf{O}$ be a union of all object classes such that 
$\vert \bigcup \textbf{O} \vert  = n$.
Let $\textbf{P}$ and $\textbf{F}$ be the given sets of physical properties' labels and a set of functional properties' labels respectively.
By abuse of notation, each physical and functional property is identified with its label.
For each physical property $P \in \textbf{P}$ as well as for a functional property $F \in \textbf{F}$, sensory data is acquired from each object instance $o \in \bigcup \textbf{O}$.
Let $P_n$ and $F_n$ represent sets of $n$ number of extracted sensory values from $n$ number of object instances for a physical property $P \in \mathbf{P}$ and a functional property $F \in \mathbf{F}$ respectively    .

\subsubsection{Sub-categorization - From Continuous to Discrete}
\label{subsubsec:subcategory}
The sub-categorization process is performed to form (more intuitive) qualitative measures to represent the degree with which a property is reflected by an object instance.
It is the first step in creating symbolic knowledge about object classes where the symbols representing the qualitative measures of a physical or a functional property reflected in an object instance are generated unsupervisedly by a clustering mechanism.
A qualitative measure of a physical property is referred to as a physical quality and that of a functional property as a functional quality.

In this process, $P_n$ and $F_n$ representing measurements of a physical property $P \in \mathbf{P}$ and a functional property $F \in \mathbf{F}$ respectively extracted from $n$ number of object instances is categorized into a given number of discrete clusters $\eta$ using a clustering algorithm.
Let $\nabla_P$ and $\nabla_F$ be partitions of the sets $P_n$ and $F_n$ after performing clustering on them. 
Let $P_\eta$ and $F_\eta$ be the sets of labels, expressing physical qualities and functional qualities, generated for a physical property $P \in \textbf{P}$ and a functional property $F \in \textbf{F}$ respectively.
Given the label for a property, the quality labels are generated by combining a property label $P$ and a cluster label (created by the clustering algorithm).
For instance, the quality labels for a property $size$ are represented as $\{ size\_1, size\_2, size\_3, size\_4 \}$. 
At the end of the sub-categorization process, the clusters are mapped to the generated symbolic labels for qualitative measures.%

Note that the number of clusters essentially describes the granularity with which each property can qualitatively be represented. 
The higher number of clusters suggest that an object is described in a finer detail which may obstruct the selection of a substitute since it may not be possible to find a substitute which is similar to a missing tool down to the finer details. 
For example, in $ size = \{ small, \text{ }medium, \text{ }big, \text{ }bigger\} $, \textit{size} is a physical property and \textit{small, medium, big, bigger} are its physical qualities.
The semantic terms given above are meant for the readers to understand the qualitative measures of the properties.

\subsubsection{Attribution - Object Instance Knowledge}
\label{subsubsec: attribution}

The attribution process generates knowledge about each object instance by aggregating all the physical and functional qualities assigned to the object instance by the \textit{sub-categorization} step.
In other terms, the knowledge about an instance consists of the physical as well as functional qualities reflected in the instance.
Let $\textbf{P}_\eta$ and $\textbf{F}_\eta$ be the families of sets containing the physical quality labels $P_\eta$ and the functional quality labels $F_\eta$ for each physical property $P \in \mathbf{P}$ and functional property $F \in \mathbf{F}$ respectively.
Thus, each object instance $o \in \bigcup \mathbf{O}$ is represented as a set of all the physical as well as functional qualities attributed to it which are expressed by a symbol $holds$ as:
$ holds \subset \bigcup \textbf{O}  \times  ( \textbf{P}_\eta  \cup \textbf{F}_\eta) $
For example, knowledge about the instance $plate_{1}$ of a \textit{plate} class can be given as,
$holds(plate_{1}, medium)$, $holds(plate_{1}, harder)$, $holds(plate_{1}, can\_support)$ where \textit{medium} is a physical quality of \textit{size} property, \textit{harder} is a physical quality of \textit{rigidity} property and \textit{can\_support} is a functional quality of \textit{support} property.

\subsubsection{Conceptualization - Knowledge about Objects}
\label{subsubsec: concept_object}

The conceptualization process aggregates the knowledge about all the instances of an object class.
The aggregated knowledge is regarded as conceptual knowledge about an object class.

Let $\textbf{O}_{KB}$ be a knowledge base about object classes where each object class $O \in \textbf{O}$.
Given the knowledge about all the instances of an object class $O$, in the conceptualization process, the knowledge about the object class $O_{K} \in \textbf{O}_{KB}$ is expressed as a set of tuples consisting of a physical or a functional quality and its proportion (membership) value in the object class. %
A tuple is expressed as $\langle O, t, m \rangle$ where $ t \in \textbf{P}_\eta  \cup \textbf{F}_\eta $ and 
a proportion value $m$ is calculated using the following membership function:
$m = P(holds(o, t) | o \in O) $. %
The proportion value allows to model the intra-class variations in the objects.

For example, knowledge about object class \textit{table} can be expressed as:
\{$\langle$\textit{plate, harder, 0.6}$\rangle$, $\langle$\textit{plate, light\_weight, 0.75}$\rangle$, 
$\langle$\textit{plate, less\_hollow, 0.67}$\rangle$, $\langle$\textit{plate, hollow, 0.33}$\rangle$, 
$\langle$\textit{plate, more\_support, 0.71}$\rangle$\}, 
where the numbers indicate that, for instance, physical quality \textit{harder} was observed in 60\% instances of object class \textit{plate}.
At the end of the conceptualization process, conceptual knowledge about an object class is created which is represented in a symbolic fuzzy form and grounded into the human-generated or machine-generated data about the properties of objects.
The knowledge about objects is then used to determine a substitute from the existing objects in the environment.

\subsubsection{Conceptualization - Knowledge about Functional Properties}
\label{subsubsec:function_model}

In addition to conceptual knowledge about objects, \textit{Conceptualization} process also creates knowledge about functional quality, termed as a function model, by associating the occurrence of physical qualities in an object instance with the occurrence of a functional quality in the instance and aggregating the result of such concurrent occurrences. 
The role of a functional model is discussed later in the section \ref{subsubsec:relevant}.
Given the knowledge about the object instances, a function model $f_d$ of a functional quality $f \in \textbf{F}_\eta$ is expressed as a set of tuples containing a functional quality $f \in \textbf{F}_\eta$, a physical quality $p \in \textbf{P}_\eta$ and a proportion value $d$.
A tuple is represented as $\langle f, p, d \rangle$ where $f \in \textbf{F}_\eta, p \in \textbf{P}_\eta $ and a proportion value $d$ is computed as, $d = P(holds(o, p)|holds(o, f))$
For example, a function model for a functional quality \textit{more\_support} is given as, \{ $\langle$\textit{more\_support, harder,0.8}$\rangle$, $\langle$\textit{more\_support, softer, 0.2}$\rangle$ where the number indicates that, for instance, functional quality $more\_support$ and a physical quality $harder$ co-occurred in the knowledge about the object instances $80$\% of the time.

\subsection{Reasoner}
\label{subsec:reasoner}
\begin{figure}
  \centering
    \includegraphics[width=0.99\textwidth]{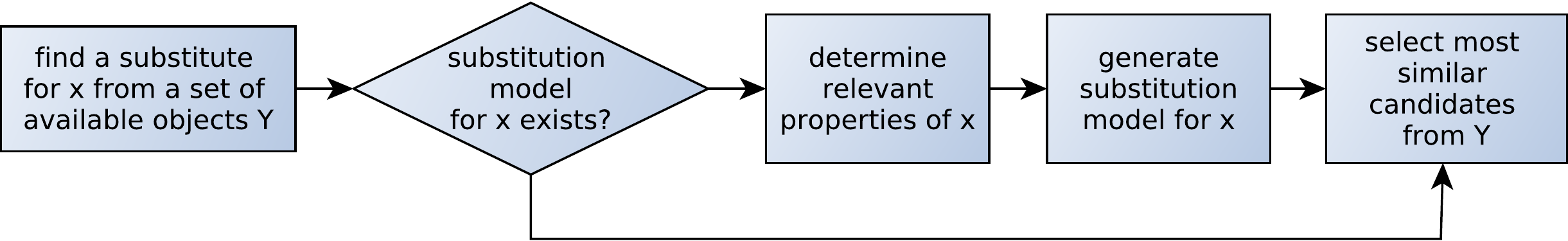}
  \caption{A typical process flow to determine a substitute for a missing tool from the available objects.} %
 \label{fig:reasoner}
\end{figure}

Fig. \ref{fig:reasoner} illustrates a process flow consisting of the primary operations involved in determining a substitute.
The flow offers an approximated aerial view for the prototypical model of ERSATZ.
When ERSATZ is queried to find a substitute for a missing tool \textit{x} from the set of available objects \textit{Y} the system checks if the substitution model for \textit{x} exists in the knowledge base.
If the substitution model does not exist, then the reasoner computes the relevant functional and physical properties of the queried tool.

\subsubsection{Representative Models}
\label{subsubsec:representative}
A representative physical model and a representative functional model of an object consists of the physical or functional qualities, respectively, that are regarded as representative qualities of the object class, while the qualities which do not fall under representative qualities are regarded as exceptional or uncommon qualities.

Let $O \in \textbf{O}$ be an object class of a missing tool and let $\theta$ is a \textit{representative model threshold} which qualifies a physical or a functional quality as stereotypical or representative to the object class $O$.
$ O_{rp} $ is called as a \textit{representative physical model} of an object class $O$ such that $ O_{rp} = \{ p : implies(O, p) \geq \theta, p \in \textbf{P}_\eta \}$
and $ O_{rf} $ is called as a \textit{representative functional model} of an object $O$ such that $ O_{rf} = \{ f : implies(O, f) \geq \theta, f \in \textbf{F}_\eta \}$.
Similarly, let $f_d$ be a function model of functional quality $f$, then $f_{rp} $ is called as a \textit{representative physical model} of a functional quality $f$ such that 
$ f_{rp} = \{ p : implies(f, p) \geq \theta,p \in \textbf{P}_\eta \}$

\subsubsection{Relevant Qualities}
\label{subsubsec:relevant}

Due to the abstract nature of an image schema and by extension a corresponding functional property, it can subsume various purposes of objects, for example,  
a functional property \textit{support} which can subsume the purposes \textit{place on, sit on} and \textit{serve on} of the a table, a chair and a tray respectively.
It is suggested in ~\cite{Baber2003_5} that a certain assemblage of physical properties are essential prerequisites to enable a functional property.
Thus, it can be assumed that by knowing the relevance of one functional property can help identify the relevant physical properties of different objects which are used for different purposes.

The relevance of a representative functional quality is decided by examining whether the physical characterization of the function model of the representative functional quality of a tool are in a close proximity to the physical characterization of a representative physical model of the tool.
The close proximity between a functional quality and the object class of the tool is determined using Jaccard Index.
Jaccard Index determines a similarity and dissimilarity between the two sets A and B where the similarity is calculated by dividing the magnitude of the intersection of A and B by the magnitude of the union of A and B. 

Let $O_{rp}$ and $f_{rp}$ be the representative physical models of an object class $O$ of the missing tool and of a function model $f_d$ of a representative functional quality $f \in \textbf{F}_\eta$ of the object class $O$ respectively. 
Let $\phi$ be a \textit{Minimum Similarity Tolerance} threshold for similarity.
Then, Jaccard Index of $O_{rp}$ and $f_{rp}$ is computed as:
$ J(O_{rp}, f_{rp}) = \frac{\vert O_{rp} \cap f_{rp} \vert}{\vert O_{rp} \cup f_{rp} \vert} $.
A representative functional quality $f$ of an object class $O$ is regarded as relevant if $J(O_{rp}, f_{rp}) > \phi$. 
Let $O_{F'}$ be a set of all relevant functional qualities of an object class $O$.
Let $f_{rp}$ be a representative physical model of a function model $f_d$ of a relevant functional quality $f \in O_{F'}$.
Let  $O_{rp}$ be a representative physical model of $O$.
Then, the relevant physical qualities of an object class $O$, expressed by a set  $O_{P'} = (O_{rp} \cap f_{rp})$.

\subsubsection{Reasoning about a Substitute}
\label{subsubsec:substitute}
Let $O^{\mu} \in \mathbf{O}$ be an object class of a missing tool and let $O^{\beta} \in \mathbf{O}$ be an object class of a possible candidate for a substitute. 
Let $O^{\mu}_{P'}$ be a set of relevant physical qualities of $O^{\mu}$ and let $ O^{\beta}_{rp} $ be a representative physical model of $O^{\beta}$.
Let $\phi$ be a \textit{Minimum Similarity Tolerance} threshold for similarity.
The substitutability of a candidate is determined by measuring the similarity between $O^{\mu}_{P'}$ and $ O^{\beta}_{rp} $ using Jaccard's Index. 
$O^{\beta}$ is termed as a substitute, expressed as $O^{\beta+}$, if $J(O^{\mu}_{P'}, O^{\beta}_{rp}) > \phi$, else it is regarded as not a substitute and expressed as $O^{\beta-}$.
Given the set of relevant physical qualities $O^{\mu}_{P'}$, the set of relevant functional qualities $O^{\mu}_{F'}$ and a positive substitute $O^{\beta+}$, and a negative substitute $O^{\beta-}$, a \textit{substitution model} of $O^{\mu}$ is expressed as a tuple:
$ \langle O^{\mu}_{P'}, O^{\mu}_{F'}, O^{\beta+}, O^{\beta-} \rangle $.
The knowledge about object $O_{\mu} \in \mathbf{O}$ is then extended in $\mathbf{O}_{KB}$ to accommodate its substitution model.%

\section{Experimental Evaluation}
\label{sec:ex_eval}
\begin{figure}[h!]
	\centering
	\subfigure[Human expert selection distributions]{\label{fig:toolsub:heat_expert}\includegraphics[height=5.5cm]{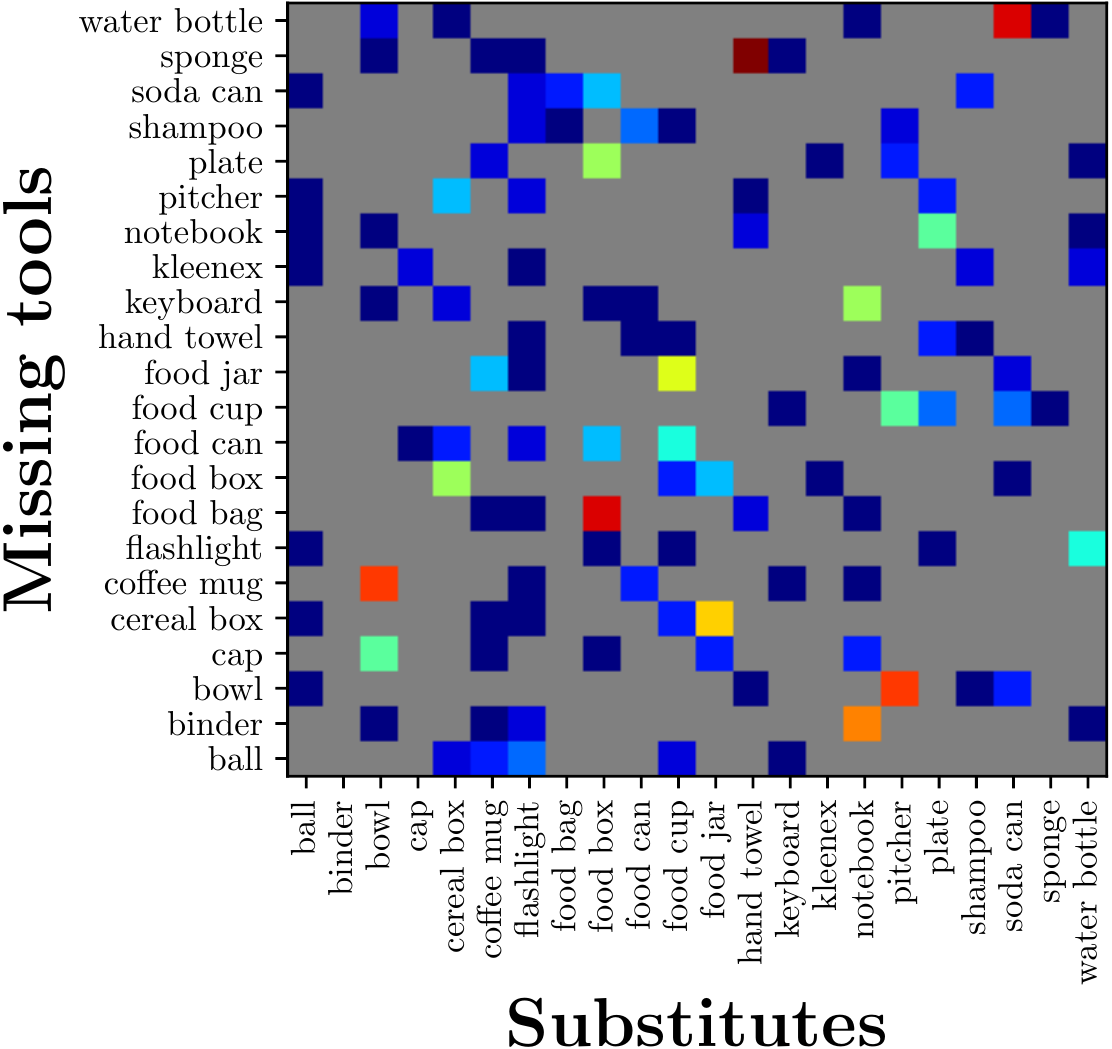}}
	\subfigure[ERSATZ selections with similarity to the missing tool]{\label{fig:toolsub:heat_ersatz}\includegraphics[height=5.55cm]{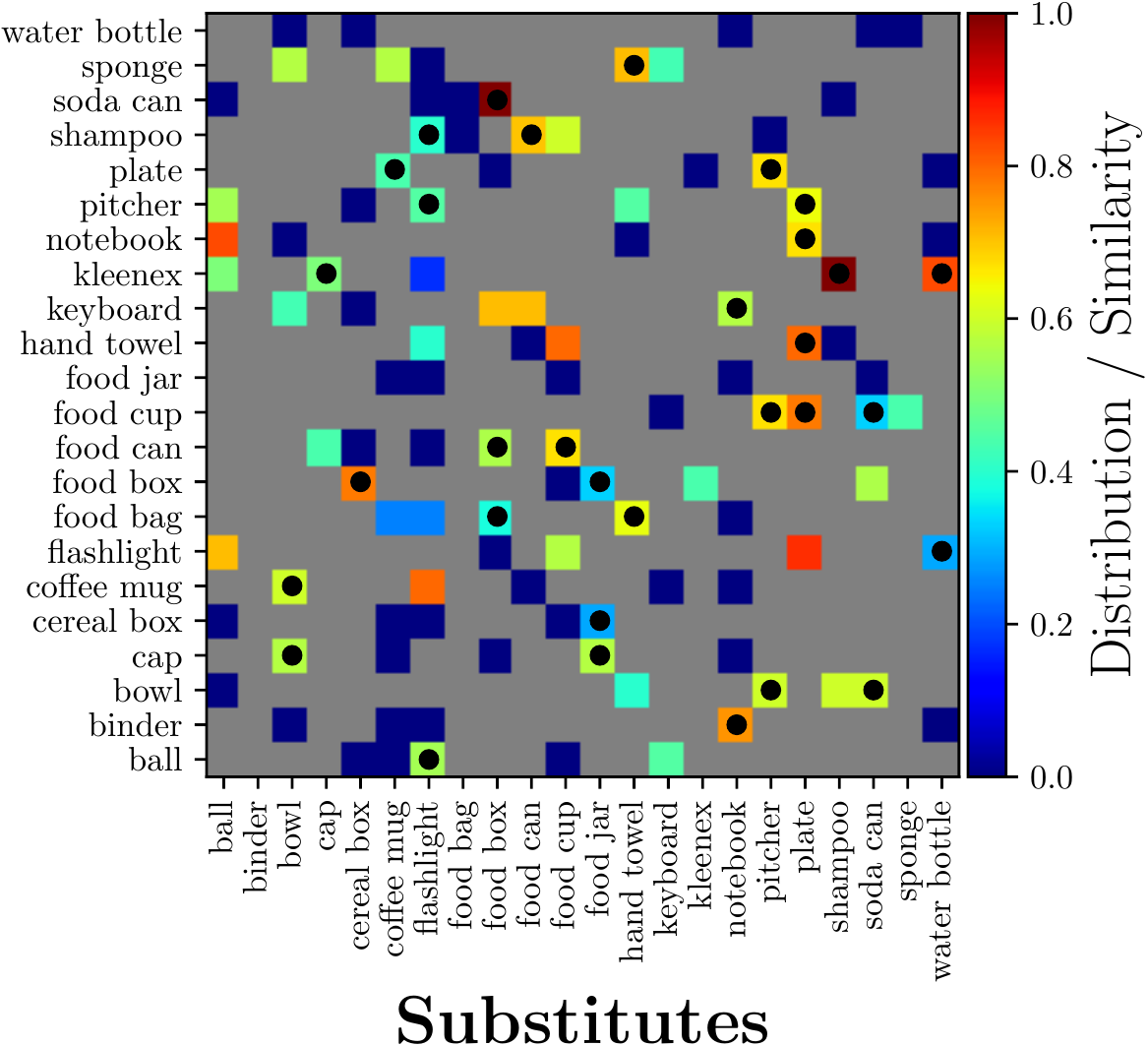}}
	\caption{Substitution results w.r.t. human expert selection \emph{distribution} and ERSATZ \emph{similarity} responses. Note that, gray cells correspond to object categories which are not available in the respective query, cells marked with \tikzdrawcircle[black, fill=black]{2.5pt} represents substitutes selected by experts and ERSATZ.}
	\label{fig:ersatz_query}
\end{figure}

The objective of the experimental evaluation of ERSATZ is 
to validate the suitability of the substitutes computed by ERSATZ by comparing the results with that of human experts.
For the experimental evaluation, we used the images from the Washington Dataset \cite{Lai2011} to generate human-based and machine-based properties.
Around $22$ object categories were selected and for each category, we selected random images from all the given instances of the category leading up to total of $692$ images. 
Table \ref{tab:datasize} illustrates the number of images selected from each category.
For the experiment, we generated $22$ queries based on $22$ object categories.
Each query consisted of a missing tool and $5$ randomly selected objects from which a substitute was to be selected.
We gave $22$ queries, %
to $14$ human experts and asked them to select a substitute in each query. 
The distribution of the human selections for each scenario is illustrated in Fig. \ref{fig:toolsub:heat_expert}.
Similarly, the queries were run on ERSATZ with the following (heuristically determined) optimal values of the target parameters: i) Number of machine-generated properties is set to $4$ (Sec. \ref{subsec:know_acqui}), ii) Number of clusters to $4$ (Sec. \ref{subsubsec:subcategory}),  iii) Representative threshold (Sec. \ref{subsubsec:representative}) and Minimum Similarity Tolerance (Sec. \ref{subsubsec:relevant}) to $0.35$.

\begin{table}%
	\caption{Number of scans (\#) per category ($\Sigma\# = 692$) of the Washington RGBD dataset~\cite{Lai2011}.}
	\label{tab:datasize}
	\centering
	\footnotesize
	\setlength{\tabcolsep}{0.05em}
	\begin{tabular}{ll|llllllllllllllllllllll}
		\footnotesize
		\rotatebox[origin=l]{90}{Category} & \rotatebox[origin=l]{90}{label} &
		 \rotatebox[origin=l]{90}{ball} & \rotatebox[origin=l]{90}{binder} & \rotatebox[origin=l]{90}{bowl} &  \rotatebox[origin=l]{90}{cap} & \rotatebox[origin=l]{90}{cereal box} & \rotatebox[origin=l]{90}{coffee mug} & \rotatebox[origin=l]{90}{flashlight} & \rotatebox[origin=l]{90}{food bag} & \rotatebox[origin=l]{90}{food box} & \rotatebox[origin=l]{90}{food can} & \rotatebox[origin=l]{90}{food cup} & \rotatebox[origin=l]{90}{food jar} & \rotatebox[origin=l]{90}{hand towel}&
		\rotatebox[origin=l]{90}{keyboard}&
		\rotatebox[origin=l]{90}{kleenex}&
		\rotatebox[origin=l]{90}{notebook}&
		\rotatebox[origin=l]{90}{pitcher}&
		\rotatebox[origin=l]{90}{plate}&
		\rotatebox[origin=l]{90}{shampoo}&
		\rotatebox[origin=l]{90}{soda can} &
		\rotatebox[origin=l]{90}{sponge}&
		\rotatebox[origin=l]{90}{water bottle}\\ 
		\rotatebox[origin=l]{90}{Inst.} &\rotatebox[origin=l]{90}{}&
		\rotatebox[origin=l]{90}{1-7}&
		\rotatebox[origin=l]{90}{1-3}&		
		\rotatebox[origin=l]{90}{1-6}&		
		\rotatebox[origin=l]{90}{1-4}&
		\rotatebox[origin=l]{90}{1-5}&
		\rotatebox[origin=l]{90}{1-8}&
		\rotatebox[origin=l]{90}{1-5}&		
		\rotatebox[origin=l]{90}{1-8}&
		\rotatebox[origin=l]{90}{1-12}&
		\rotatebox[origin=l]{90}{1-14}&
		\rotatebox[origin=l]{90}{1-5}&
		\rotatebox[origin=l]{90}{1-6}&
		\rotatebox[origin=l]{90}{1-5}&
		\rotatebox[origin=l]{90}{1-5}&
		\rotatebox[origin=l]{90}{1-5}&
		\rotatebox[origin=l]{90}{1-5}&
		\rotatebox[origin=l]{90}{1-3}&
		\rotatebox[origin=l]{90}{1-7}&
		\rotatebox[origin=l]{90}{1-6}&	
		\rotatebox[origin=l]{90}{1-6}&
		\rotatebox[origin=l]{90}{1-12}&
		\rotatebox[origin=l]{90}{1-9} \\ 
		\rotatebox[origin=l]{90}{Scans} & \rotatebox[origin=l]{90}{per Inst.} &
		\rotatebox[origin=l]{90}{5}&
		\rotatebox[origin=l]{90}{10}&		
		\rotatebox[origin=l]{90}{5}&		
		\rotatebox[origin=l]{90}{8}&
		\rotatebox[origin=l]{90}{6}&
		\rotatebox[origin=l]{90}{4}&
		\rotatebox[origin=l]{90}{6}&		
		\rotatebox[origin=l]{90}{4}&
		\rotatebox[origin=l]{90}{3}&
		\rotatebox[origin=l]{90}{2}&
		\rotatebox[origin=l]{90}{6}&
		\rotatebox[origin=l]{90}{5}&
		\rotatebox[origin=l]{90}{6}&
		\rotatebox[origin=l]{90}{6}&
		\rotatebox[origin=l]{90}{6}&
		\rotatebox[origin=l]{90}{6}&
		\rotatebox[origin=l]{90}{10}&
		\rotatebox[origin=l]{90}{5}&
		\rotatebox[origin=l]{90}{5}&	
		\rotatebox[origin=l]{90}{5}&
		\rotatebox[origin=l]{90}{3}&
		\rotatebox[origin=l]{90}{4}
		\\ \hline
		\#& &35& 30&30&32& 30& 32 &30& 32& 36& 28&30&30&30&30&30&30&30&35&30&30&36&36\\
	\end{tabular}
\end{table}

The results of both experiments were plotted as a heat map where the y-axis shows missing tools and x-axis shows the available objects illustrated in Fig. \ref{fig:ersatz_query}.
The grayed cells mean the corresponding object categories were not available in the respective query.
The cells that are marked with \tikzdrawcircle[black, fill=black]{2.5pt} represents substitutes selected by experts and ERSATZ.
Out of $22$ scenarios,
ERSATZ and the experts identified the same substitutes in $20$ scenarios ($91$\%).

\section{Future Work}
\label{sec:future_work}
The paper presents a prototypical system to determine a substitute for a missing tool using the grounded knowledge about objects.
The approach has drawn inspiration from symbol grounding, the theory of affordances and the theory of image schemas to represent the grounded knowledge and to determine a substitute.
This is an ongoing research with a focus on the following aspects.

Our immediate goal focuses on the fuzzification of the clustering method and the reasoning method to combat the migration of the data points within clusters. 
Moreover, we have derived three functional properties, namely, \textit{contain}, \textit{support}, \textit{block} from the image schemas \textit{Containment}, \textit{Support} and \textit{Blockage} respectively.
However, further investigation is needed to formalize the identification of additional functional properties to be derived from the existing image schema.
\begin{figure}[tb]
   \centering
     \includegraphics[width=0.5\textwidth]{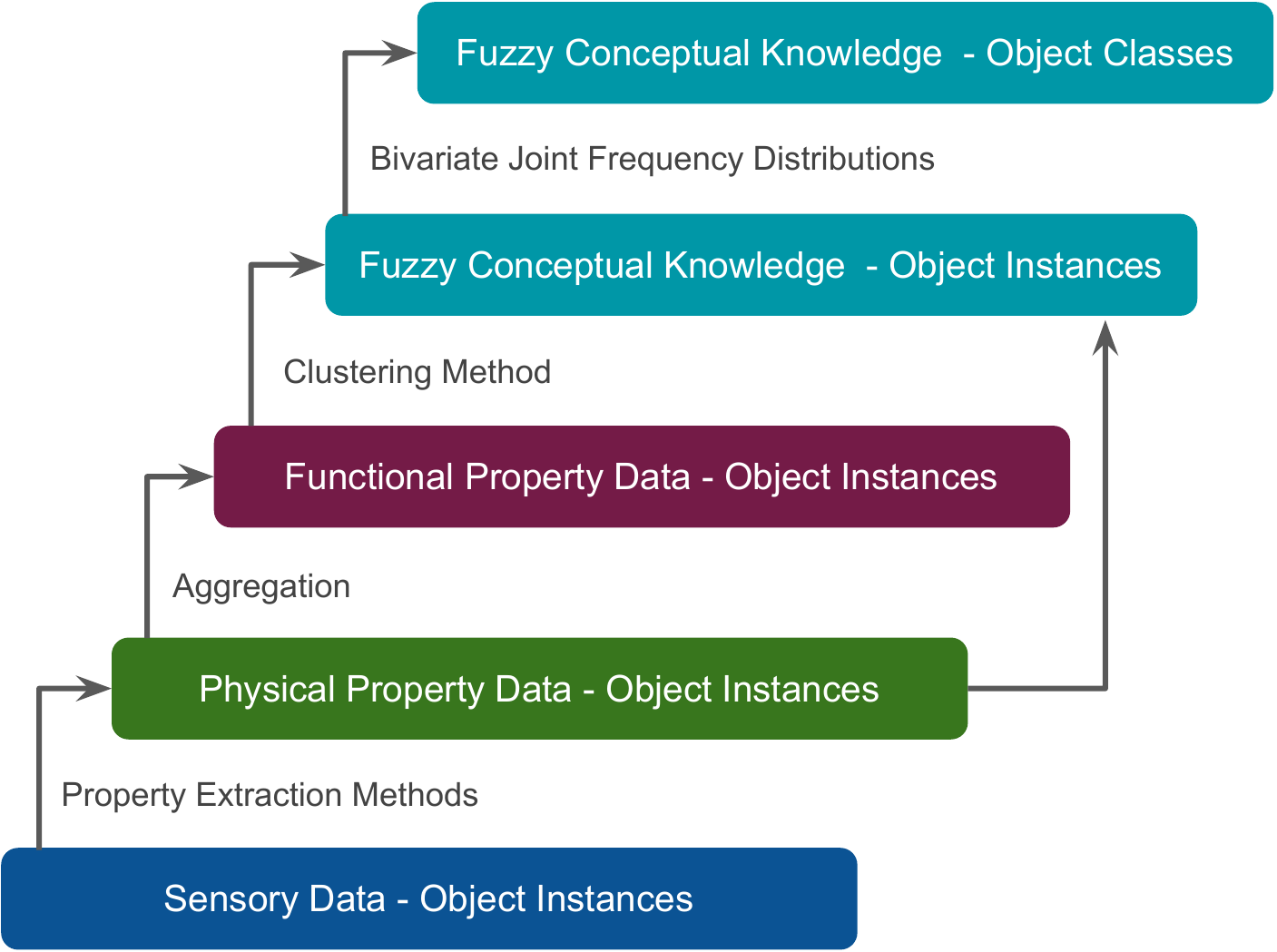}
   \caption{Multi-layered dataset to build a robot-centric grounded knowledge about objects.}
  \label{fig:dataset_approach}
\end{figure}
For robot-centric property acquisition, we are currently developing a framework that allows a robot to extract properties of individual objects and build a knowledge base in a bottom-up manner such that the knowledge about properties of objects is constructed on the basis of what is sensed (see Fig.~\ref{fig:dataset_approach}).
We have proposed the preliminary framework in \cite{jamutho2018}.

\bibliographystyle{aaai}

\end{document}